\documentclass[10pt,twocolumn,letterpaper]{article}

\usepackage{cvpr}
\usepackage{times}
\usepackage{epsfig}
\usepackage{graphicx}
\usepackage{amsmath}
\usepackage{amssymb}
\usepackage{slashbox}
\usepackage{sidecap}

\usepackage{pifont}

\usepackage{multirow}
\usepackage{colortbl}

\usepackage[pagebackref=false,breaklinks=false,letterpaper=true,colorlinks,bookmarks=false]{hyperref}

\usepackage{fancyhdr,lipsum}
\fancypagestyle{firstpage}{
  \fancyhf{}
  \fancyhead[C]{Presented at SUNw: Scene Understanding Workshop, CVPR 2016}
}
\cvprfinalcopy 

\def\cvprPaperID{729} 

\ifcvprfinal\fi
\begin{document}

\title{Improving object recognition performance by analyzing invariance \\ properties of deep neural networks}
\title{Evaluating object recognition accuracy and invariance using a large-scale controlled dataset}
\title{Enhancing accuracy and invariance of deep neural networks using a large-scale controlled object dataset}
\title{Enhancing accuracy and invariance of deep neural networks using a large-scale controlled object dataset}
\title{New insights into deep learning using a large-scale controlled object dataset}
\title{What can we learn about CNNs from a large-scale controlled object dataset?}

\title{Vanishing point detection with convolutional neural networks}



\author{Ali Borji\\
Center for Research in Computer Vision, University of Central Florida\\
\texttt{\small{aborji@crcv.ucf.edu}}
}

\maketitle

\thispagestyle{firstpage}

\section{Introduction}
\vspace{-5pt}
In a graphical perspective, a \textit{vanishing point} (VP) is a 2D point (in the image plane) which is the intersection of parallel lines in the 3D world (but not parallel to the image plane). In other words, the vanishing point is the spot to which the receding parallel lines diminish. In principle, there can be more than one vanishing point in the image.
VP can commonly be seen in fields, railroads, streets, tunnels, forest, buildings,
objects such as ladder (from looking bottom-up), etc. It is an important visual cue useful in several applications (e.g., camera calibration, 3D reconstruction, autonomous driving). 

Inspired by the finding that vanishing point (road tangent) guides driver's gaze~\cite{land1994we,borjiPAMI}, in our previous work we showed that vanishing point attracts gaze during free viewing of natural scenes as well as in visual search~\cite{borji2015vanishing}. 
We have also introduced improved saliency models using vanishing point detectors~\cite{feng2015fixation}. 
Here, we aim to predict vanishing points in naturalistic environments by training convolutional neural networks in an end-to-end manner. 

Traditionally, geometrical and structural features such as lines and corners (e.g., using Hough transform~\cite{hough1962method}) have been applied for detecting vanishing points in images. 
Here, we follow a data-driven learning approach by training two popular convolutional neural networks, Alexnet and VGG, for: 1) predicting whether a vanishing point exists in a scene (on a $n \times n$ grid map), and 2) If so, we then attempt to localize its exact location. 

\begin{figure*}
\includegraphics[width=17.5cm,height=7.5cm]{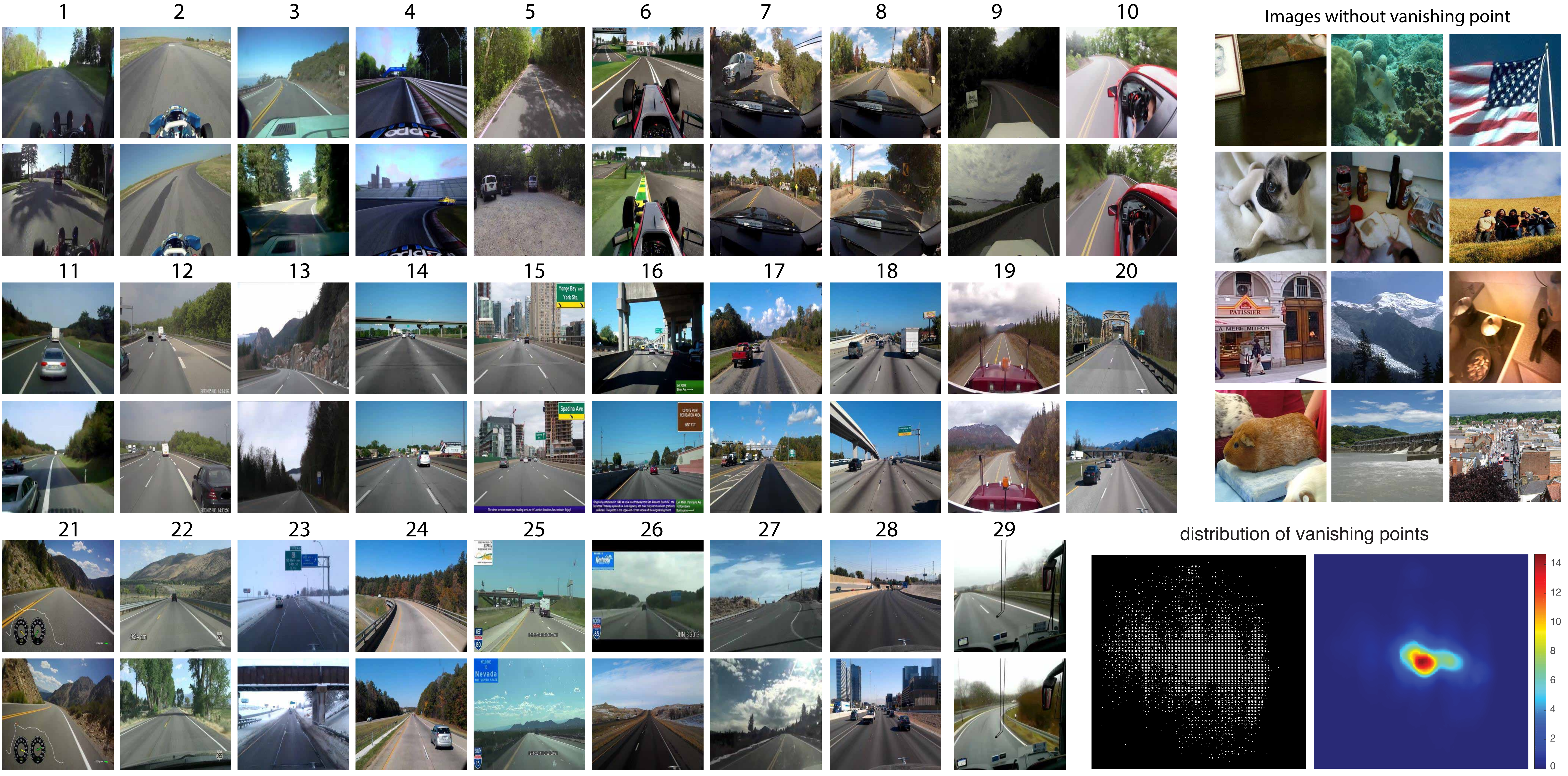}
\caption{Left: Two sample frames of each of 29 videos downloaded from YouTube. Top-right: Sample images without vanishing point used to train the vanishing point existence prediction network. Bottom-right: Average vanishing point location. Left panel shows all visited locations and the right panel shows the VP histogram.}
\label{fig:VpExamples}
\end{figure*}

\vspace{-5pt}
\section{Experiments \& Results}
\vspace{-5pt}
\subsection{Data collection}
\vspace{-5pt}
To train deep neural networks, often a large amount of data is needed. We resorted to YouTube to download videos including road trips across America (e.g., from sedan, bus, or truck dash cams), personal adventures (e.g., using shifters or motorbikes) or game playing sessions (e.g., formula one, Nascar). 
These videos have been captured in a variety of weather and ground conditions (e.g., freeway, race track, in city, inter city, snowy, rainy, sunny, mountainous, forest, vegetation). Eventually, we had 37,497 frames (resized to 300 $\times$ 300 pixels). We annotated vanishing points (1 per frame) in all videos (one annotator; the author). The grid cell containing the vanishing point has the label 1 on a $n \times n$  grid map ($n$= [10, 20, 30]). Some example frames of 29 YouTube videos are shown in Figure~\ref{fig:VpExamples}. 

We also collected some images without VPs to train a binary classifier for VP existence prediction. A total of 32,419 images were sampled from these datasets: MIT saliency benchmark~\cite{mit-saliency-benchmark}, CAT2000 dataset~\cite{borji2015cat2000},
Caltech 256~\cite{griffin2007caltech}, 15 category dataset~\cite{torralba2006contextual} except the street and highway categories, MS COCO~\cite{lin2014microsoft}, and Imagenet~\cite{deng2009imagenet}.


\subsection{Vanishing point existence prediction}
\vspace{-5pt}
After training networks for 20 epochs over 63,916 images (34,497 with VP and 29,419 without VP), we obtained 98.9\% VP existence prediction accuracy using the Alexnet network and 99.73\% using the VGG network over the test set (6,000 images; 3,000 with VP). 

%

\subsection{Vanishing point localization}
\vspace{-5pt}
Alexnet and VGG networks were trained to map a scene into the VP location which is one of the $p$ classes ($p$ = 100, 400, or 900; linearized $n \times n$ grids). Thus, there are $p$ neurons in the output layers. We used 33,000 frames for training and the remaining 4,497 frames for testing. The network training was stopped after 40 epochs. 

\begin{figure*}
\includegraphics[width=17.5cm,height=4.5cm]{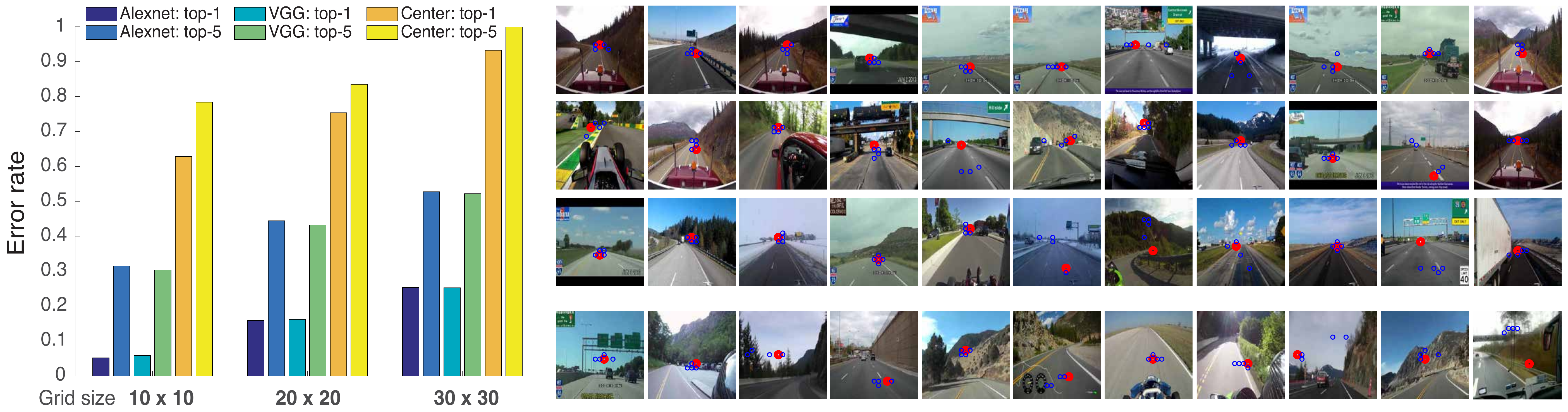}
\caption{Left: Error rates of deep models for VP detection. Top-right: Sample images where our model is able to accurately locate the VP in five tries. Red circle is the top-1 prediction and blue ones are the next top-4. Bottom-right: Failure examples of our model.}
\label{fig:VPResults}
\end{figure*}

Results are shown in Figure~\ref{fig:VPResults}. We achieved the lowest top-5 error rate of 5.1\% over 10 $\times$ 10, 15.9\% over 20 $\times$ 20, and 25.3\% over 30 $\times$ 30 grid sizes using the VGG network. It means that the probability of hitting within 15 pixels of the true VP location in 5 guesses is about 85\% (over a 20 $\times$ 20 grid on a 300 $\times$ 300 image). Our results are nearly the same using both networks. Figure~\ref{fig:VPResults} (right) shows some success and failure cases of the Alexnet for VP localization.

Since vanishing point usually happens at the image center (See Figure~\ref{fig:VpExamples}, bottom-right), we devised two baseline predictors to further evaluate our method. The first one is the most frequent grid location ([x y] in training data) denoted as the `Top-1 center' and the second one is the five most frequent locations ([x y], [x-1 y], [x y-1], [x+1 y], [x y+1]; all set to one, the rest are zero) denoted as `Top-5 center'. These models perform well above chance (16.5\% accuracy using Top-1 center vs. 0.25\% chance over a 20 $\times$ 20 grid) but are well below the deep learning performance (deep learning Top-1 accuracy is about 57\%). 


We also compare our model with two vanishing point detection algorithms from the literature. The first one is a method by 
Ko{\v{s}}eck{\'a} and Zhang~\cite{kovsecka2002video} and the second one is the classic Hough transform~\cite{hough1962method}. These two algorithms score 15.6\% and 35\%, respectively in detecting the vanishing point on a  20 $\times$ 20 map (Top-1 accuracy) which are much lower than our results using CNNs. 



\begin{figure*}[t]
  \centering
  \includegraphics[width=17.5cm,height=7.3cm]{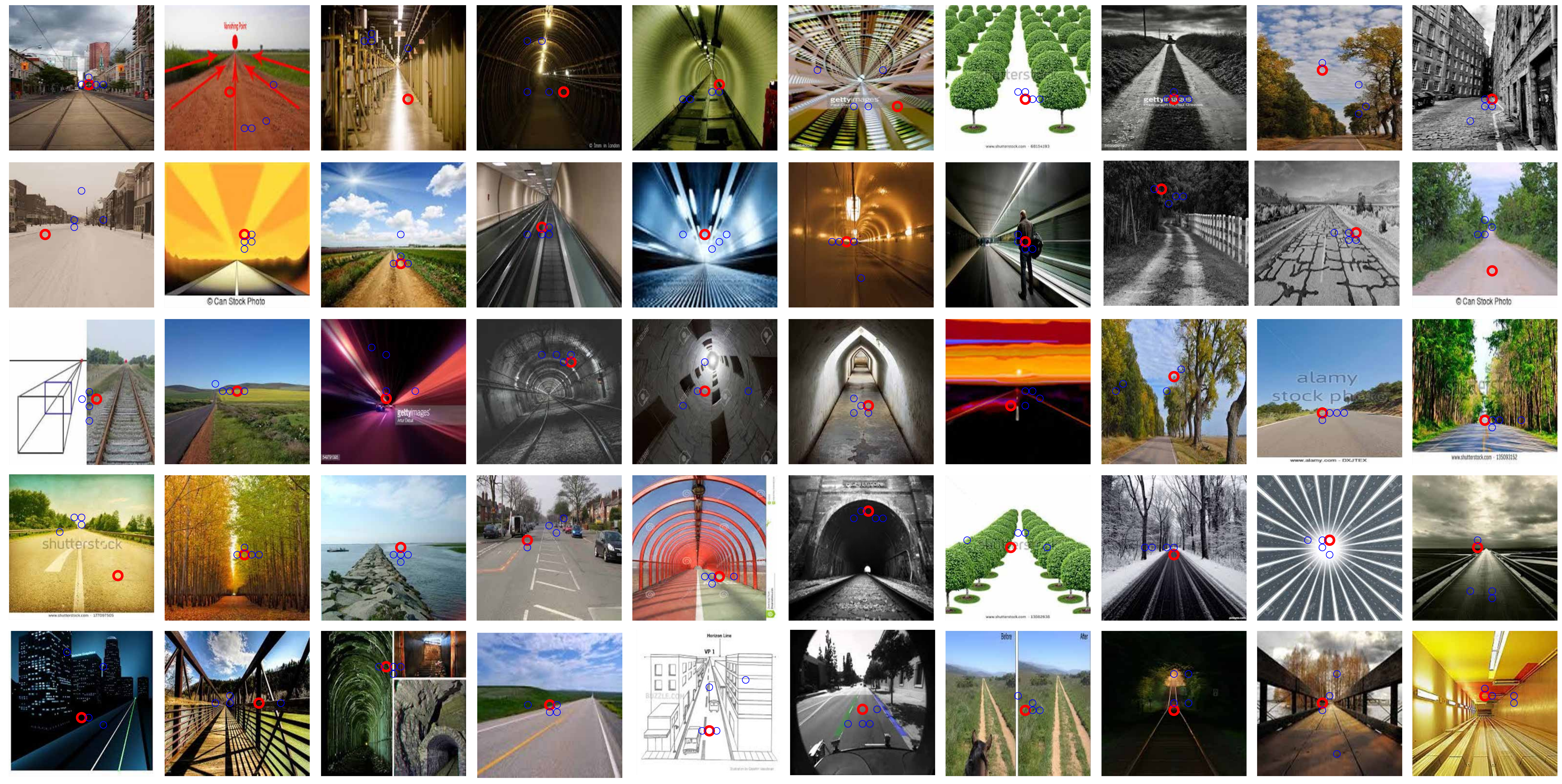}
  \caption{Performance of our vanishing point detector on arbitrary images containing vanishing points. The largest red circle is the first detection. Other four detections are shown in blue.}
  \vspace{5pt}
\label{fig:VPFailures}
\end{figure*}

To assess the generalization power of our approach in detecting vanishing points in arbitrary natural scenes, we experimented with pictures of buildings, tunnels, sketches and fields shown in Figure~\ref{fig:VPFailures}. Although our model (VGG) has not been explicitly trained on these images, it successfully finds VPs in some of them. It fails on some other unseen examples (e.g., sketches). Augmenting our dataset with more images of these kinds could help overcome this shortcoming. Another way to improve performance would be through data augmentation (i.e., adding jittered, cropped, noisy, and blurry versions of input images).



%

\vspace{-5pt}
\section{Discussion}
\vspace{-5pt}
We proposed a method for vanishing point detection based on convolutional neural networks that does well on road scenes but is not very effective on arbitrary images. We will consider collecting a larger image dataset with variety of scenes including vanishing points and more recent deep learning architectures to improve accuracy. Extension of this approach to videos is another interesting future direction. Our dataset is freely available at: http://crcv.ucf.edu/people/faculty/Borji/code.php

\vspace{2pt}

\textbf{Acknowledgments:} We wish to thank NVIDIA for their generous
donation of the GPU used in this study.


\end{document}